\documentclass[conference]{IEEEtran}
\IEEEoverridecommandlockouts
\usepackage{cite}
\usepackage{amsmath,amssymb,amsfonts}
\usepackage{algorithmic}
\usepackage{textcomp}
\usepackage{xcolor}
\usepackage{booktabs}
\usepackage{graphicx}
\usepackage{subcaption}
\usepackage{hyperref}
\usepackage{xurl}
\usepackage{balance}
\usepackage{placeins}

\def\BibTeX{{\rm B\kern-.05em{\sc i\kern-.025em b}\kern-.08em
    T\kern-.1667em\lower.7ex\hbox{E}\kern-.125emX}}

\usepackage{threeparttable}
\usepackage{multirow}
    
\begin{document}

\title{Evaluating Zone-Guided Front Extraction for Glacier Calving-Front Delineation in SAR Imagery}

\author{
\IEEEauthorblockN{Chhaya Kulkarni}
\IEEEauthorblockA{
Department of Computer and Information Sciences\\
Towson University\\
ckulkarni@towson.edu
}
\and
\IEEEauthorblockN{Emam Hossain}
\IEEEauthorblockA{
Department of Information Systems\\
University of Maryland, Baltimore County\\
emamh1@umbc.edu
}
}

\maketitle

\begin{abstract}
Automatic calving-front delineation from synthetic aperture radar imagery is challenging because the front is a thin and often ambiguous boundary between glacier ice, ocean, and surrounding rock or terrain. The CAlving Fronts and where to Find thEm (CaFFe) dataset provides both binary calving-front masks and broader semantic zone masks, making it possible to study whether zone-level supervision can support front recovery. In this paper, we compare direct front prediction with zone-guided front extraction using U-Net, DeepLabV3+, and SegFormer-B0 under the same bounding-box-cropped CaFFe setting. In the direct setting, models predict the binary calving-front mask. In the zone-guided setting, the model first predicts four semantic zone classes, and the front is then extracted from the predicted glacier–ocean boundary. We evaluate both zone-level and front-level performance, include a ground-truth-zone boundary check, and examine lightweight test-time adaptation on sensor-specific and glacier-specific subsets. The results show that zone labels contain useful front-boundary information: extracting the front from ground-truth zones gives the lowest mean distance error. However, fronts extracted from model-predicted zones remain weak, even when zone segmentation scores are moderate. Test-time adaptation also does not consistently improve zone-guided front recovery. These results indicate that zone segmentation performance should not be treated as a substitute for front-level evaluation and that effective use of zone labels may require boundary-aware training, label fusion, or explicit front supervision.

\end{abstract}


\section{Introduction}

Marine-terminating glaciers lose mass partly through calving, where ice breaks away from the glacier front into the ocean. Tracking the position of the calving front over time is important for monitoring glacier change, frontal ablation, and ice-sheet mass loss \cite{rignot2011ice,mouginot2019forty,imbie2020mass}. Manual delineation of calving fronts in satellite imagery is time-consuming, and automated methods remain challenging because the front is often thin, irregular, and visually ambiguous, especially in synthetic aperture radar (SAR) images.

SAR imagery is useful for glacier monitoring because it can be acquired through cloud cover and during polar night \cite{gourmelon2022calving,holzmann2021glacier}. However, SAR-based calving-front delineation is difficult due to speckle, sensor differences, variable spatial resolution, seasonal surface changes, and the presence of ice mélange near the front \cite{wu2023amd,gourmelon2026comparison}.

The CaFFe dataset provides a useful benchmark for studying this problem because it includes SAR images together with both binary calving-front annotations and broader semantic zone masks \cite{gourmelon2022calving}. The front annotations represent the calving front as a thin binary line, while the zone masks label broader scene regions such as ocean, rock outcrops, glacier area, and no-information regions. This structure makes it possible to compare direct front prediction with a zone-guided strategy, where a model first predicts scene regions and the front is later extracted from the glacier-ocean boundary.

Most prior work on CaFFe has focused on improving direct front delineation using specialized architectures, self-supervised SAR representations, or temporal information \cite{wu2023amd,gourmelon2026comparison,gourmelon2025ssl4sar,dreier2025multi}. In contrast, less attention has been given to whether the available zone labels can help standard segmentation models recover calving fronts. This is a useful question because zone labels provide denser supervision than sparse front masks. At the same time, accurate zone segmentation may not be sufficient for accurate front delineation if the predicted glacier-ocean boundary is not locally precise.

This paper evaluates that question through a controlled comparison of direct front prediction and zone-guided front extraction on CaFFe. We use three common segmentation models: U-Net, DeepLabV3+, and SegFormer-B0. In the direct setting, the models are trained to predict the binary front mask. In the zone-guided setting, the models are trained to predict four-class zone masks, and the front is extracted from the predicted glacier-ocean boundary. We evaluate both zone-level and front-level performance to determine whether region-level segmentation accuracy translates into useful front recovery.

The contributions of this work are as follows:
\begin{itemize}
\item We compare direct front prediction and zone-guided front extraction using U-Net, DeepLabV3+, and SegFormer-B0 under the same bounding-box-cropped CaFFe setting.
\item We evaluate whether predicted zone masks can support calving-front recovery by extracting the glacier-ocean boundary from model-predicted zones.
\item We include a ground-truth-zone boundary check to determine whether the zone labels themselves contain useful front-boundary information.
\item We report zone-level and front-level metrics separately, showing that zone segmentation performance does not necessarily imply accurate front delineation.
\item We examine lightweight test-time adaptation on sensor-specific and glacier-specific subsets to assess whether inference-time adjustments improve direct-front and zone-guided predictions.
\end{itemize}

\section{Related Work}

Gourmelon et al. introduced CaFFe as a benchmark for calving-front extraction from multi-sensor SAR imagery \cite{gourmelon2022calving}. Its binary labels represent the front as a thin line, while its four-class zone masks describe glacier, ocean or ice mélange, rock, and no-information regions. Because a small displacement can sharply reduce pixel overlap, CaFFe studies commonly report mean distance error (MDE) together with F1. A recent comparison also shows that current automated systems remain below human annotator performance \cite{gourmelon2026comparison}.

Several methods improve front delineation by preserving local boundary detail while incorporating broader context. Herrmann et al. examined alternative label formulations with nnU-Net, showing that supervision design can affect sparse-boundary extraction \cite{herrmann2023out}. AMD-HookNet and AMD-HookNet++ combine high-resolution and contextual features through specialized multi-branch or hybrid CNN--transformer designs \cite{wu2023amd,wu2025amd}. Other work uses self-supervised SAR representations, change information, or multiple acquisition times to improve robustness \cite{gourmelon2025ssl4sar,zhao2025cisnet,dreier2025multi}.

These studies mainly target stronger direct-front models. Our work instead asks a narrower question: whether CaFFe's denser zone supervision can support front recovery with standard segmentation architectures. We therefore evaluate zone quality and the recovered front separately, since accurate region segmentation may still yield an imprecise glacier-ocean boundary.

\section{Methodology}
\begin{figure*}[!h]
    \centering
    \includegraphics[width=\textwidth]{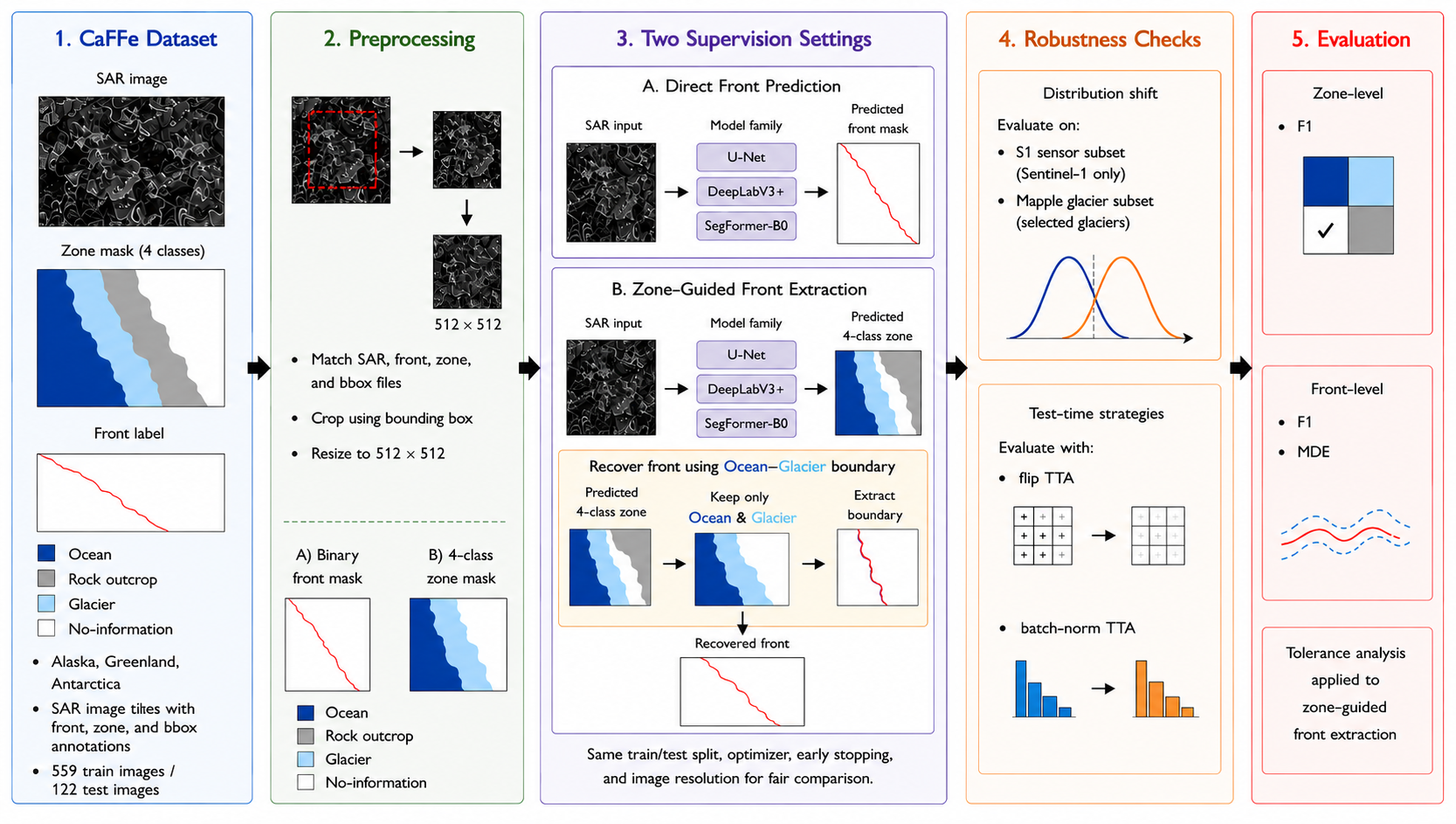}
    \caption{Overview of the experimental pipeline. The same bounding-box-cropped SAR inputs are evaluated under two supervision settings: direct front prediction and zone-guided front extraction. In the zone-guided branch, the calving front is recovered from the predicted glacier–ocean boundary. Subset analyses on the Sentinel-1 and Mapple subsets apply flip and BatchNorm test-time adaptation. Colors in the zone masks denote the four semantic classes (ocean, rock outcrop, glacier, no-information).}
    \label{fig:overview}
\end{figure*}

We compare direct front prediction with zone-guided front extraction using the same SAR inputs, crops, image size, and train/test split. Figure~\ref{fig:overview} summarizes the pipeline.

\subsection{Dataset and Preprocessing}
CaFFe contains 681 SAR images acquired from 1995--2020, each paired with a binary front mask, a four-class zone mask, and a bounding box around the dynamic front region \cite{gourmelon2022calving}. We map the zone classes to no-information, rock/outcrop, ocean, and glacier. The same bounding box is applied to the SAR image and both masks so that all inputs remain spatially aligned. The resulting variable-size crops are resized to $512 \times 512$ pixels. Bilinear interpolation is used for continuous SAR intensities, whereas nearest-neighbor interpolation preserves discrete mask labels. The prepared split contains 559 training and 122 test images. The bounding-box-cropped setting focuses the analysis on the dynamic frontal region rather than large areas unrelated to calving-front localization. Applying the same crop to the SAR image, front mask, and zone mask ensures that the image pixels and their corresponding labels remain aligned. Resizing all crops to a common resolution also enables consistent batching and comparison across models.

\subsection{Model Selection}
We compare U-Net, DeepLabV3+, and SegFormer-B0, representing encoder-decoder CNN, atrous multi-scale CNN, and hierarchical transformer segmentation, respectively\cite{ronneberger2015u,chen2018encoder,xie2021segformer}.

All models used the same split, $512 \times 512$ inputs, batch size 2, 100-epoch limit, and early stopping after 10 epochs without validation macro-F1 improvement. Four-class zone models used cross-entropy loss. U-Net and DeepLabV3+ used Adam with learning rates of $10^{-4}$ and $5\times10^{-4}$, respectively; SegFormer-B0 used AdamW with learning rate $5\times10^{-5}$ and weight decay $10^{-4}$.

% \subsection{Model Selection}
% We evaluate U-Net, DeepLabV3+, and SegFormer-B0 to cover three common segmentation families \cite{ronneberger2015u,chen2018encoder,xie2021segformer}. U-Net uses encoder--decoder skip connections to preserve spatial detail; DeepLabV3+ combines atrous multi-scale context with decoder refinement; and SegFormer-B0 uses hierarchical transformer features with a lightweight decoder. This selection allows the two supervision settings to be compared without introducing a glacier-specific architecture.

% For a controlled comparison, all models used the same train/test split, $512 \times 512$ inputs, a batch size of 2, a maximum of 100 epochs, and early stopping after 10 epochs without improvement in validation macro-F1. Cross-entropy loss was used for four-class zone prediction. Architecture-specific optimizers and learning rates were used: Adam with a learning rate of $1 \times 10^{-4}$ for U-Net, Adam with a learning rate of $5 \times 10^{-4}$ for DeepLabV3+, and AdamW with a learning rate of $5 \times 10^{-5}$ and weight decay of $1 \times 10^{-4}$ for SegFormer-B0.

\subsection{Direct Front Prediction}
In the direct setting, the cropped SAR image is mapped directly to a binary calving-front mask. This optimizes the final output of interest, but the target is highly sparse because the front is represented as a thin line.

\subsection{Zone-Guided Front Extraction}
In the zone-guided setting, the same input is mapped to the four-class zone mask. At inference, only glacier and ocean predictions are retained, and their shared boundary is extracted as the estimated front. The extracted line is then evaluated against the binary front annotation. This tests whether denser region-level supervision can recover a precise sparse boundary.

\subsection{Ground-Truth-Zone Boundary Check}
To determine whether the zone labels themselves encode usable front information, we also extract the glacier-ocean boundary from the ground-truth zone masks. This diagnostic is not a trained-model result; it separates errors caused by zone prediction from those introduced by the boundary-extraction procedure.

\subsection{Test-Time Adaptation}
We test lightweight inference-time strategies on the Sentinel-1 (S1) and Mapple subsets, which showed the largest per-group SegFormer-B0 errors among sensor and glacier groups. For SegFormer-B0, horizontal flip test-time augmentation averages predictions from the original image and its flipped version. For DeepLabV3+, we also test BatchNorm adaptation where applicable; it updates normalization statistics using unlabeled test images \cite{nado2020evaluating,schneider2020improving}; related test-time methods also adapt parameters \cite{wang2020tent}.

\subsection{Evaluation Metrics}
Zone segmentation is evaluated using macro F1 across the four classes. Front delineation is evaluated using front F1 and mean distance error (MDE) in pixels. F1 requires exact pixel overlap, whereas MDE measures spatial separation between predicted and annotated lines and is therefore informative for thin boundaries \cite{gourmelon2022calving,gourmelon2026comparison}. All front-level distances are measured in the resized $512 \times 512$ image coordinate system. The same evaluation procedure is applied to direct-front predictions, fronts recovered from predicted zones, and the ground-truth-zone diagnostic to maintain consistency. Tolerance-based F1 is used only as a diagnostic to determine whether poor zone-guided overlap is caused by small shifts or by fragmented and incomplete recovery.

\section{Results and Discussion}

\subsection{Direct Front Prediction and Zone-Guided Extraction}
Table~\ref{tab:direct_zone} compares direct and zone-guided prediction; the ground-truth-zone row is diagnostic rather than a trained result. 
Among trained models, direct-front SegFormer-B0 performs best (F1 0.0939; MDE 35.79). SegFormer-B0 also gives the highest zone F1 (0.6808), but its extracted front remains weak (F1 0.0058; MDE 66.37). Thus, good region segmentation does not ensure accurate local glacier-ocean boundaries.

% \begin{table}[t]
% \centering
% \caption{Comparison of direct front prediction and zone-guided front extraction.}
% \scriptsize
% \label{tab:direct_zone}
% \resizebox{\columnwidth}{!}{%
% \begin{tabular}{l l c c c}
% \toprule
% \textbf{Model} & \textbf{Setting} & \textbf{Zone F1}$\uparrow$ & \textbf{Front F1}$\uparrow$ & \textbf{MDE}$\downarrow$ \\
% \midrule
% U-Net & Direct front & -- & 0.0510 & 79.32 \\
% U-Net & Zone-guided & 0.5814 & 0.0055 & 64.65 \\
% DeepLabV3+ & Direct front & -- & 0.0427 & 65.47 \\
% DeepLabV3+ & Zone-guided & 0.3642 & 0.0060 & 90.06 \\
% SegFormer-B0 & Direct front & -- & 0.0939 & 35.79 \\
% SegFormer-B0 & Zone-guided & 0.6808 & 0.0058 & 66.37 \\
% Ground-truth zones & Diagnostic & -- & 0.1478 & 21.01 \\
% \bottomrule
% \end{tabular}%
% }
% \vspace{1mm}
% \begin{minipage}{0.97\columnwidth}
% \footnotesize
% \textit{Note:} Zone F1 is not applicable for direct-front models because they predict binary front masks rather than zone masks.
% \end{minipage}
% \end{table}

\begin{table}[!h]
\centering
\caption{Direct front prediction vs.\ zone-guided front extraction on the CaFFe test set. The ground-truth-zone row (italic) is a diagnostic upper reference, not a trained model.}
\label{tab:direct_zone}
\begin{threeparttable}
\scriptsize
\resizebox{\columnwidth}{!}{%
\begin{tabular}{l l c c c}
\toprule
\textbf{Model} & \textbf{Setting} & \textbf{Zone F1}$\uparrow$ & \textbf{Front F1}$\uparrow$ & \textbf{MDE}$\downarrow$ \\
\midrule
\multirow{2}{*}{U-Net} & Direct front & -- & 0.0510 & 79.32 \\
 & Zone-guided & 0.5814 & 0.0055 & 64.65 \\
\addlinespace
\multirow{2}{*}{DeepLabV3+} & Direct front & -- & 0.0427 & 65.47 \\
 & Zone-guided & 0.3642 & 0.0060 & 90.06 \\
\addlinespace
\multirow{2}{*}{SegFormer-B0} & Direct front & -- & \textbf{0.0939} & \textbf{35.79} \\
 & Zone-guided & \textbf{0.6808} & 0.0058 & 66.37 \\
\midrule
\textit{Ground-truth zones} & \textit{Diagnostic} & -- & \textit{0.1478} & \textit{21.01} \\
\bottomrule
\end{tabular}%
}
\begin{tablenotes}[flushleft]
\footnotesize
\item Zone F1 is undefined for direct-front models, which predict binary front masks rather than zone masks (``--'').
\end{tablenotes}
\end{threeparttable}
\end{table}

\subsection{Ground-Truth-Zone Boundary Check}
The ground-truth-zone check clarifies the zone-guided results. Extracting the front from ground-truth zones increases front F1 to 0.1478 and reduces MDE to 21.01 pixels, the lowest among all settings. This confirms that the zone labels contain useful boundary information. The poor performance of predicted-zone fronts therefore reflects the inability of standard models to preserve the glacier-ocean boundary precisely enough, rather than a lack of information in the zone labels. Thus, zone labels are useful in principle, but simple zone prediction followed by boundary extraction remains insufficient for accurate calving-front delineation.
% The ground-truth-zone boundary check helps interpret the zone-guided results. When the front is extracted from the ground-truth zones, front F1 increases to 0.1478 and MDE decreases to 21.01 pixels. This is the lowest MDE among all evaluated settings. This result shows that the zone labels contain useful front-boundary information. The weak performance of the model-predicted zones should therefore not be interpreted as evidence that zone labels are uninformative. Rather, the bottleneck is that the trained models do not predict the glacier-ocean boundary precisely enough for reliable front extraction. In other words, the zone labels are useful in principle, but simple zone prediction followed by boundary extraction is not sufficient for accurate calving-front delineation.

\subsection{Tolerance-Based Analysis of Zone-Guided Fronts}
% Tolerance-based F1 improves only slightly for predicted-zone fronts but substantially for ground-truth-zone fronts. The errors are therefore not merely small shifts; predicted fronts are often fragmented or incomplete.
Tolerance-based F1 was evaluated at 0, 1, 2, 3, 5, and 10 pixels. Predicted-zone F1 remained below 0.01 for all models, peaking at 0.0095 for SegFormer-B0 at three pixels. Thus, weak zone-guided recovery is not explained by minor boundary shifts alone.

\subsection{Test-Time Adaptation on S1 and Mapple Subsets}
Table~\ref{tab:tta} reports test-time results on the S1 and Mapple subsets. Flip TTA improves direct-front SegFormer-B0 MDE on both subsets, but not its zone-guided outputs. DeepLabV3+ BatchNorm adaptation also provides no reliable front-level gains and often worsens MDE. Lightweight test-time changes therefore do not resolve the boundary-precision limitation of zone-guided extraction.

\begin{table}[t]
\centering
\caption{Test-time adaptation on the worst-case Sentinel-1 (S1) and Mapple subsets. ``Front'' denotes models trained on binary front masks; ``Zone'' denotes models trained on zone masks with the front extracted from the predicted glacier-ocean boundary.}
\label{tab:tta}
\begin{threeparttable}
\scriptsize
\resizebox{\columnwidth}{!}{%
\begin{tabular}{l l c c c c c}
\toprule
\textbf{Model} & \textbf{Target} & \textbf{Set} & \textbf{Strategy} & \textbf{Zone F1}$\uparrow$ & \textbf{Front F1}$\uparrow$ & \textbf{MDE}$\downarrow$ \\
\midrule
\multirow{8}{*}{Seg-B0} & \multirow{4}{*}{Front} & \multirow{2}{*}{S1} & None & -- & 0.1109 & 43.82 \\
 & & & Flip & -- & 0.1150 & \textbf{38.13} \\
 & & \multirow{2}{*}{Mapple} & None & -- & 0.1105 & 36.15 \\
 & & & Flip & -- & 0.1054 & \textbf{33.98} \\
\cmidrule(l){2-7}
 & \multirow{4}{*}{Zone} & \multirow{2}{*}{S1} & None & 0.5134 & 0.0018 & \textbf{97.67} \\
 & & & Flip & 0.5174 & 0.0022 & 102.83 \\
 & & \multirow{2}{*}{Mapple} & None & 0.7154 & 0.0051 & \textbf{70.38} \\
 & & & Flip & 0.6703 & 0.0041 & 80.93 \\
\midrule
\multirow{10}{*}{DLV3+} & \multirow{4}{*}{Front} & \multirow{2}{*}{S1} & None & -- & 0.0737 & \textbf{64.45} \\
 & & & BN & -- & 0.0853 & 102.36 \\
 & & \multirow{2}{*}{Mapple} & None & -- & 0.0755 & \textbf{52.32} \\
 & & & BN & -- & 0.0627 & 70.30 \\
\cmidrule(l){2-7}
 & \multirow{6}{*}{Zone} & \multirow{3}{*}{S1} & None & 0.2884 & 0.0044 & \textbf{76.76} \\
 & & & Flip & 0.3164 & 0.0037 & 83.34 \\
 & & & BN & 0.3208 & 0.0027 & 90.29 \\
 & & \multirow{3}{*}{Mapple} & None & 0.4486 & 0.0071 & 90.61 \\
 & & & Flip & 0.4791 & 0.0067 & \textbf{86.45} \\
 & & & BN & 0.4989 & 0.0045 & 92.22 \\
\bottomrule
\end{tabular}%
}
\begin{tablenotes}[flushleft]
\footnotesize
\item Seg-B0: SegFormer-B0; DLV3+: DeepLabV3+; BN: BatchNorm adaptation. Zone F1 is undefined for front-trained models (``--''). Flip TTA is applied to SegFormer-B0 and BN adaptation to DeepLabV3+, matching each model's applicable mechanism.
\end{tablenotes}
\end{threeparttable}
\end{table}

\section{Conclusion and Future Work}

Zone labels contain useful calving-front information, but standard zone-prediction models do not preserve the glacier-ocean boundary precisely enough for reliable extraction. Direct-front SegFormer-B0 performs best, while the ground-truth-zone check suggests that boundary-aware use of zone supervision remains promising. Flip TTA and BatchNorm adaptation provide no consistent gains. Because this study evaluates standard segmentation models with simple boundary extraction, the results should be viewed as a controlled comparison rather than a state-of-the-art benchmark. Future work should combine zone and front supervision using boundary-aware losses or explicit glacier-ocean constraints, together with temporal information and SAR-specific pretraining to improve robustness.

% Zone labels contain useful calving-front information, but standard zone-prediction models do not preserve the glacier-ocean boundary accurately enough for reliable extraction. Direct-front SegFormer-B0 outperforms the zone-guided trained models, while the ground-truth-zone check shows that boundary-aware use of zone supervision remains promising. Flip TTA and BatchNorm adaptation provide no consistent gains. 

% A limitation of this study is that it evaluates standard segmentation models and simple boundary extraction, rather than proposing a new boundary-aware architecture. The results should therefore be interpreted as a controlled analysis of direct-front and zone-guided supervision, not as a state-of-the-art comparison.

% Future work should combine zone and front supervision using boundary-aware losses or explicit glacier-ocean constraints. Temporal information and SAR-specific pretraining may further improve robustness across sensors, glaciers, and seasons.

\section*{Acknowledgment}
This work is supported by iHARP: NSF HDR Institute for Harnessing Data and Model Revolution in the Polar Regions (Award\# 2118285). The views expressed in this work do not necessarily reflect the policies of the NSF, and endorsement by the Federal Government should not be inferred.

\balance
\bibliographystyle{IEEEtran}
\bibliography{refs}

\end{document}